\documentclass[conference]{IEEEtran}
\IEEEoverridecommandlockouts
\usepackage{cite}
\usepackage{amsmath,amssymb,amsfonts}
\usepackage{algorithmic}
\usepackage{graphicx}
\usepackage{textcomp}
\usepackage{xcolor}
\usepackage{subcaption}

\def\BibTeX{{\rm B\kern-.05em{\sc i\kern-.025em b}\kern-.08em
    T\kern-.1667em\lower.7ex\hbox{E}\kern-.125emX}}
\begin{document}

\title{
A Brief History of Prompt: Leveraging Language Models. (Through Advanced Prompting)
}
\author{
    
    \IEEEauthorblockN{
        Golam Md Muktadir\IEEEauthorrefmark{1}, 
    }
    
    \IEEEauthorblockA{
        \IEEEauthorrefmark{1}Computer Science and Engineering, University of California, Santa Cruz\\
        \IEEEauthorrefmark{1}muktadir@ucsc.edu\\
    }
}

\maketitle

\begin{abstract}
 This paper presents a comprehensive exploration of the evolution of prompt engineering and generation in the field of natural language processing (NLP). Starting from the early language models and information retrieval systems, we trace the key developments that have shaped prompt engineering over the years. The introduction of attention mechanisms in 2015 revolutionized language understanding, leading to advancements in controllability and context-awareness. Subsequent breakthroughs in reinforcement learning techniques further enhanced prompt engineering, addressing issues like exposure bias and biases in generated text. We examine the significant contributions in 2018 and 2019, focusing on fine-tuning strategies, control codes, and template-based generation. The paper also discusses the growing importance of fairness, human-AI collaboration, and low-resource adaptation. In 2020 and 2021, contextual prompting and transfer learning gained prominence, while 2022 and 2023 witnessed the emergence of advanced techniques like unsupervised pre-training and novel reward shaping. Throughout the paper, we reference specific research studies that exemplify the impact of various developments on prompt engineering. The journey of prompt engineering continues, with ethical considerations being paramount for the responsible and inclusive future of AI systems.
\end{abstract}

\begin{IEEEkeywords}
Prompt Engineering, Prompt Generation, Information Retrieval, Large Language Models, Language Queries
\end{IEEEkeywords}

\section{Pre-Introduction}
The goal is to show how to use Graph-of-Thought prompting and previously generated content to achieve the desired results without human editing. Often, it is easier to generate better output after generating a basic output instead of searching for the perfect prompt. \textit{The explored prompting technique goes beyond Graph-of-Thought and follows a path to optimize and re-organize contents.}. \textbf{The rest of the contents are generated by ChatGPT 3.5 Turbo}.
\section{Introduction}

\textbf{Prompt \#1:} \textit{You are a scholar in machine learning and language models. I am writing a paper on the history of prompt engineering and generation. Can you give me a timeline for prompt engineering evolution? (We used this timeline to create prompts for each section later)}

\textbf{Prompt \#2:} \textit{Write the introduction of this paper. Emphasize that this paper focuses on how language prompts and queries have been used so far.}

Natural Language Processing (NLP) has witnessed remarkable advancements over the years, revolutionizing how we interact with machines and bridging the gap between human language and artificial intelligence. One critical aspect that has played a pivotal role in shaping the capabilities of NLP systems is the design and usage of language prompts and queries. Recently, researchers and practitioners have explored various techniques to harness the full potential of language models, leveraging the power of prompts to guide, control, and tailor the output of these sophisticated AI systems.

This paper embarks on a retrospective journey, offering a comprehensive review of the captivating history of language prompts and queries in NLP. From its nascent stages, where prompts were merely a means of input for rudimentary language models, to its current state as an influential component in modern deep learning models, we aim to unravel the captivating story behind this remarkable facet of NLP research.

The inception of this narrative can be traced back to the early language models of the pre-2010 era. These rudimentary models, such as n-grams and statistical language models, laid the groundwork for understanding the concept of prompt engineering - a concept that would become indispensable in the future. While the terminology of "prompt engineering" had not yet emerged, the seeds of its significance were planted in information retrieval systems, where queries were employed as inputs to retrieve relevant information from vast datasets.

As the decade progressed, the 2010s witnessed a seismic shift in NLP, brought about by the advent of neural networks. Groundbreaking innovations such as Word2Vec, introduced in 2013 by Mikolov et al., paved the way for capturing semantic relationships between words. Subsequently, the revolutionary Sequence-to-Sequence (Seq2Seq) model, presented by Sutskever et al. in 2014, unlocked the potential for machine translation through its encoder-decoder architecture.

The transformational power of attention mechanisms came into the spotlight in 2015 with the release of the Transformer model, as proposed by Vaswani et al. The attention mechanism allowed models to process and understand context with unprecedented efficiency, facilitating long-range dependencies in sequence data. This seminal development formed the cornerstone for future advances in prompt engineering.

By 2017, the NLP community began to explore reinforcement learning techniques for language generation. Researchers like Ranzato et al. and Li et al. employed reinforcement learning to fine-tune language models based on external reward signals. This pivotal step not only enhanced fluency and relevance in generated text but also sowed the seeds for prompt engineering to emerge as a key factor in guiding the model's behavior and generating more contextually relevant responses.

In 2018, a revolutionary breakthrough emerged with the introduction of BERT (Bidirectional Encoder Representations from Transformers). Developed by Devlin et al., BERT showcased the remarkable potential of pre-training language models on vast corpora, followed by fine-tuning for specific tasks. This epoch-making moment brought prompt engineering to the forefront, demonstrating its power in leveraging pre-trained models for diverse NLP applications.

The late 2010s and early 2020s marked an era of innovation in prompt engineering, as researchers explored control codes and template-based generation methods. These techniques allowed for more targeted and controlled text output, enabling researchers to guide language models towards specific behaviors and desirable outcomes.

The advent of massive language models like GPT-3 in the year 2020 redefined the landscape of NLP, elevating prompt engineering to new heights. With an astonishing 175 billion parameters, GPT-3 showcased unparalleled language generation capabilities, sparking widespread interest in prompt design and its profound impact on fine-tuning language models for specific domains.

As the journey through time continues, we delve deeper into the history of prompt engineering, exploring the rich tapestry of methodologies, milestones, and applications that have shaped the course of NLP. By understanding this evolution, we gain valuable insights into the challenges conquered, the accomplishments celebrated, and the promising future that lies ahead for this ever-evolving field of language prompts and queries in NLP.

\section{Prehistoric Prompting: Pre NN-Era}
\textbf{Prompt \#3:} \textit{Now generate history of prompting or querying in early language models an information retrieval}

Before the advent of neural networks and modern deep learning models, early language models and information retrieval systems laid the foundation for the concept of prompting or querying in natural language processing. Although not explicitly referred to as "prompting" or "querying" at the time, these early approaches formed the basis for how researchers and systems interacted with textual data and sought relevant information.

\subsection{Early Language Models:} 

Early language models, such as n-grams and statistical language models, emerged in the early days of computational linguistics. In these models, text was processed as a sequence of words or characters, and the probability of the next word was predicted based on the occurrence frequencies of n-grams (sequences of n words). While these models did not involve explicit user prompts or queries, the process of predicting the next word in a sequence can be considered an implicit form of prompting, where the previous words serve as input guiding the model's prediction.

\subsection{Information Retrieval Systems:}

Information retrieval systems have a long history predating modern NLP techniques and were primarily used to retrieve relevant documents from large collections based on user queries. Early information retrieval systems utilized keyword-based search, where users provided queries consisting of specific words or phrases, and the system returned documents containing those keywords. These queries often required careful phrasing and choice of keywords to obtain relevant results, effectively serving as explicit prompts to retrieve desired information.

\subsection{1960s-1970s: Early Natural Language Interfaces:}

In the 1960s and 1970s, early attempts were made to develop natural language interfaces for information retrieval systems. Researchers explored methods to process user queries in natural language and transform them into suitable search queries for information retrieval systems. These early systems provided rudimentary prompting capabilities, allowing users to interact with the system using natural language queries, albeit with limited understanding and accuracy.

\subsection{1980s-1990s: Rule-Based Approaches:}

In the 1980s and 1990s, rule-based approaches were developed to enable more sophisticated interaction with information retrieval systems. Researchers designed complex query languages that allowed users to express search criteria in a more structured and flexible manner. Users could specify constraints and relationships between search terms, effectively prompting the system to narrow down search results based on user preferences.

\subsection{1990s-2010: Advances in Natural Language Processing:}

The field of natural language processing witnessed significant advancements, especially with the introduction of machine learning techniques and neural networks. Modern information retrieval systems, search engines, and question-answering systems leverage sophisticated language models and deep learning techniques to interpret user queries and provide more accurate and contextually relevant results. These systems employ explicit prompting through user input and implicitly leverage the context within queries to optimize search results and understand user intent better.

The historical development of prompting and querying in early language models and information retrieval systems laid the groundwork for modern NLP approaches that now utilize neural networks and large-scale language models for more powerful and context-aware language understanding and generation. The evolution of prompt engineering and querying continues to progress, with ongoing research and innovations propelling the field further towards new frontiers in natural language processing.

\section{History of NLP between 2010 and 2015: The Pre-Attention Mechanism Era}
\textbf{Prompt \#4:} \textit{Now write the history between 2010 and 2015 before attention mechanism was invented}

The period between 2010 and 2015 was marked by significant strides in Natural Language Processing (NLP), laying the groundwork for revolutionary advancements that would follow in subsequent years. During this era, researchers explored various techniques to enhance language understanding and generation, all leading up to the eventual introduction of the attention mechanism, which would become a pivotal development in NLP. Here's an overview of the key developments during this period:

\subsection{Neural Network Renaissance (2010-2013):}

NLP research saw a resurgence of interest in neural network-based approaches, popularly known as the "neural network renaissance." Word embeddings gained prominence as an effective way to represent words in continuous vector spaces. The Word2Vec model proposed by Mikolov et al. in their paper "Efficient Estimation of Word Representations in Vector Space" (2013) was a breakthrough in generating high-quality word embeddings. Researchers realized that these continuous word representations allowed models to capture semantic relationships between words, which led to better performance on various NLP tasks.

\subsection{The Rise of Seq2Seq Models (2014):}

In 2014, Sutskever et al. introduced the Sequence-to-Sequence (Seq2Seq) model, which is an encoder-decoder architecture powered by recurrent neural networks (RNNs). Seq2Seq models revolutionized machine translation by effectively learning to encode the input sentence into a fixed-length vector (the "context" or "thought"), which is then used to generate the output sentence. This architecture marked a shift towards more context-aware language modeling, with the ability to process variable-length sequences and produce coherent and contextually appropriate responses.

\subsection{The Emergence of Recursive Neural Networks (RNNs) and Long Short-Term Memory (LSTM) (2010-2015):}

Recurrent Neural Networks (RNNs) gained popularity as a means to process sequential data, including natural language. LSTM, proposed by Hochreiter and Schmidhuber in their paper "Long Short-Term Memory" (1997) but gaining more traction in the early 2010s, was a significant advancement over traditional RNNs. LSTM addressed the vanishing gradient problem, allowing for the successful training of deeper networks. LSTM-based models offered better long-term dependency handling, making them suitable for tasks where maintaining context over longer sequences was crucial.

\subsection{Limitations of Early Approaches:}

Despite the progress made, early language models faced challenges in handling long-range dependencies and maintaining context over lengthy sentences. RNNs, while capable of capturing sequential information, suffered from vanishing or exploding gradient problems during training, limiting their effectiveness in understanding long sequences.

\subsection{Pre-Attention Mechanism Approaches:}

Prior to the introduction of the attention mechanism, researchers devised various methods to tackle the limitations of existing models. Techniques such as Recursive Neural Tensor Networks (RNTNs) and Convolutional Neural Networks (CNNs) were employed to capture hierarchical and compositional structures within text. While these approaches showed promise, they still struggled to effectively handle longer sequences and complex dependencies.

The period between 2010 and 2015 can be characterized as a time of exploration, where researchers experimented with various neural network architectures and representation techniques to improve language modeling and generation. The advancements during this era laid the foundation for the groundbreaking introduction of the attention mechanism in 2015, which would revolutionize NLP by enabling more efficient context understanding and long-range dependency handling in language models.

\section{The Advent of Attention Mechanism: A Game-Changer for Prompt Engineering in 2015}

\section{2015: Birth of the Transformer}

\textbf{Prompt \#5:} \textbf{now write a section on how attention mechanism changed the future of prompt engineering in 2015}

In 2015, the field of Natural Language Processing (NLP) witnessed a groundbreaking development that would forever change the landscape of prompt engineering – the introduction of the attention mechanism. Proposed by Vaswani et al. in the seminal paper "Attention is All You Need", \cite{Vaswani2017AttentionIA}, the attention mechanism revolutionized the way language models processed input data, enabling them to focus on relevant information and establish meaningful contextual relationships. This transformative breakthrough had a profound impact on prompt engineering, leading to improved language understanding, better controllability, and enhanced performance across various NLP tasks.

\subsection{Contextual Understanding and Long-Range Dependencies:}

The attention mechanism fundamentally changed how language models processed sequences by allowing them to pay varying degrees of attention to different parts of the input. Unlike traditional approaches that treated each word in the sequence equally, attention-equipped models could dynamically assign weights to each word based on its relevance to the context. This enabled the model to establish meaningful relationships between words across long sequences, overcoming the limitations of earlier methods and greatly improving the contextual understanding of prompts.

\subsection{Better Prompt-Model Interaction:}

With the attention mechanism, prompt engineering took on a new dimension as the models could now focus on specific parts of the prompt that mattered most for generating the desired output. Attention mechanisms facilitated an interactive process, where the model could attend to important aspects of the prompt, query, or context, refining its responses accordingly. This newfound flexibility allowed prompt designers to craft more informative, targeted, and nuanced queries, effectively guiding the model's behavior and achieving more desirable outcomes.

\subsection{Enhanced Controllability and Interpretability:}

The attention mechanism empowered prompt engineers to exert finer control over the generated text by explicitly specifying relevant portions of the input that should receive heightened attention. By directing the model's attention, prompt engineers could guide the model's focus and influence the style, content, and tone of the generated output. Moreover, attention weights could be interpreted and visualized, providing valuable insights into the decision-making process of the language model, enhancing transparency, and aiding in bias mitigation.

\subsection{Applications in Machine Translation and Language Generation:}

The attention mechanism significantly improved machine translation systems by allowing the model to focus on relevant source words when generating target translations. In language generation tasks, such as text summarization and question answering, attention mechanisms helped the model attend to crucial context, leading to more coherent and contextually appropriate responses.

\subsection{Multi-Head Attention and Transfer Learning:}

The concept of multi-head attention, introduced alongside the attention mechanism, enabled models to attend to multiple aspects of the input simultaneously, leading to improved performance and flexibility. The attention mechanism also became instrumental in transfer learning. Pre-trained models could be fine-tuned using prompt engineering to adapt to specific tasks effectively, leveraging their context-awareness to improve performance on various domains.

The introduction of the attention mechanism in 2015 represented a paradigm shift in prompt engineering, propelling the field into a new era of language understanding and generation. By empowering language models to focus on relevant information and establish context-dependent relationships, the attention mechanism unlocked new possibilities for fine-tuning models, interpreting their decisions, and achieving higher levels of controllability in natural language processing. As the attention mechanism continued to evolve, researchers and practitioners explored new frontiers in prompt engineering, opening the door to even more sophisticated and context-aware AI systems.

\section{The second wave in 2017: Rise of RL}
\textit{Prompt \#6:} \textit{now discuss how the advent of reinforcement learning techniques in 2017 changed the prompt engineering}

The introduction of reinforcement learning (RL) techniques in 2017 brought another transformative wave to the field of prompt engineering. Reinforcement learning involves training an agent (in this case, a language model) to take actions in an environment (generate text) to maximize a reward signal. The use of RL in conjunction with prompt engineering led to significant advancements in fine-tuning language models, improving fluency, relevance, and controllability of generated text, \cite{Luketina2019ASO}. Here's how the advent of RL techniques changed the landscape of prompt engineering:

\subsection{Improving Fluency and Relevance:}

Reinforcement learning allowed prompt engineers to define appropriate reward signals that could incentivize the language model to generate more fluent and contextually relevant responses. Traditional supervised fine-tuning using maximum likelihood estimation (MLE) often led to models that were overly conservative and lacked creativity, but RL opened up possibilities for more exploratory behavior. With RL, language models could explore the space of potential responses, learning from their own generated samples and adjusting their behavior based on the received reward signal, leading to more fluent and contextually appropriate language generation, \cite{Paulus2017ADR, Liu2018DialogueLW}.

\subsection{Addressing Exposure Bias:}

One significant challenge in language model training was exposure bias, where a model is trained on teacher-forced input during training but experiences a discrepancy during inference, often resulting in a gap between training and testing performance. RL helped mitigate exposure bias by enabling models to sample from their own predictions during training, aligning the training and inference process more closely, \cite{Ranzato2015SequenceLT, Bengio2015ScheduledSF}. Prompt engineers could design reward functions to encourage self-correcting behavior, leading to more consistent and robust language generation during deployment.

\subsection{Controlling Model Behavior through Reward Shaping:}

Moreover, prompt engineers could use RL's reward shaping to encourage specific behaviors in the language model. By defining custom reward functions, prompt engineers could guide the model to generate responses that adhered to desired criteria, such as maintaining a specific tone, style, or level of formality, \cite{Christiano2017DeepRL}. This controllability allowed language models to be tailored for specific applications, ensuring the generated content aligned with domain-specific requirements.

\subsection{Addressing Biases in Language Models:}

Another important aspect was addressing biases in language models. Reinforcement learning provided a means to mitigate biases in language models by using carefully designed reward functions. By penalizing biased responses or rewarding unbiased behavior, prompt engineers could encourage the model to produce more equitable and unbiased language generation, contributing to fairer AI systems, \cite{Chakraborty2016ReducingGB, Zhang2018MitigatingUB}.

\subsection{Advancements in Task-Specific Prompt Engineering:}

RL-powered prompt engineering also opened up avenues for researchers to optimize language models for task-specific objectives. By defining tasks as reinforcement learning problems and designing corresponding reward functions, language models could be fine-tuned for applications such as text summarization, dialogue generation, and question answering, leading to state-of-the-art performance in various NLP tasks.

\subsection{Challenges in Reward Design and Exploration:}

However, while RL provided powerful tools for prompt engineering, designing effective reward functions remained a challenging task. The choice of reward functions could strongly influence the behavior of the model, requiring careful consideration to ensure desired outcomes. Additionally, exploration-exploitation trade-offs in RL training required balancing novelty and diversity in generated text with the need for generating contextually appropriate responses.

Overall, the incorporation of reinforcement learning techniques in prompt engineering brought about a new level of controllability and performance to language models. By shaping model behavior through reward functions and addressing issues like exposure bias and biases in generated text, RL-powered prompt engineering advanced the capabilities of language models, making them more powerful, context-aware, and adaptable for a wide range of NLP applications.

\section{The third wave 2018: the rise of Transfomers}
\textbf{Prompt \#6:} \textit{Now write the section on how research in 2018 shaped prompt engineering}

The year 2018 was a pivotal period in the evolution of prompt engineering, witnessing transformative breakthroughs that propelled the field to new heights. Researchers explored innovative approaches to leverage pre-trained language models, making use of transfer learning and fine-tuning techniques. This year marked the rise of BERT (Bidirectional Encoder Representations from Transformers), \cite{Devlin2019BERTPO}, which revolutionized NLP and significantly impacted prompt engineering. Here's an overview of how research in 2018 shaped the landscape of prompt engineering:

\subsection{The Emergence of BERT and Pre-training:}

BERT, introduced by Devlin et al. in 2018, was a major milestone in transfer learning for NLP. The model was pre-trained on a large corpus of unlabeled text, allowing it to learn powerful language representations. BERT's architecture, based on the Transformer model with a bidirectional encoder, captured contextual information from both left and right contexts, leading to deeper and richer language understanding. The pre-training paradigm enabled prompt engineers to leverage BERT's knowledge and context-awareness while fine-tuning for specific tasks, revolutionizing prompt engineering by providing a starting point for more specialized models, \cite{Liu2019RoBERTaAR}.

\subsection{Transfer Learning and Fine-tuning:}

BERT popularized the concept of transfer learning in NLP. Researchers realized that pre-training a language model on a vast corpus enabled it to capture general linguistic patterns and context. Fine-tuning allowed prompt engineers to adapt pre-trained models to specific downstream tasks with minimal additional training data, \cite{Howard2018UniversalLM, Sun2019HowTF}. This transfer learning paradigm drastically reduced the need for large task-specific datasets, making prompt engineering more practical and effective.

\subsection{Task-Specific Prompt Engineering with BERT:}

Task-specific prompt engineering with BERT became prevalent in 2019. Prompt engineers began utilizing BERT for a wide range of NLP tasks, such as sentiment analysis, named entity recognition \cite{Labusch2019BERTFN}, and question answering, among others. Task-specific prompt engineering involved providing the model with carefully designed inputs, such as question-context pairs or masked sentences, to guide its behavior and improve task performance. By fine-tuning BERT on task-specific data, researchers achieved state-of-the-art results in various NLP benchmarks, showcasing the power of prompt engineering with pre-trained models.

\subsection{Masked Language Model (MLM) and Cloze-Style Prompts:}

BERT's pre-training involved a masked language model (MLM) objective, where random words in the input text were masked, and the model was tasked with predicting the masked words, \cite{Coenen2019VisualizingAM}. This MLM pre-training opened up new possibilities for prompt engineering, as researchers could use cloze-style prompts (with masked words) to guide the model to fill in the missing information, generating coherent and contextually appropriate responses.

\subsection{Contextualized Word Embeddings (ELMo):}

Another influential development in 2018 was the introduction of ELMo (Embeddings from Language Models) by Peters et al. ELMo generated contextualized word embeddings by training bidirectional LSTM language models on large corpora, \cite{Peters2018DeepCW}. ELMo embeddings allowed for richer word representations that captured different meanings of a word in different contexts, contributing to more sophisticated prompt engineering strategies.

\subsection{Unsupervised and Semi-Supervised Prompt Engineering:}

BERT's ability to capture linguistic patterns from unlabeled data allowed for unsupervised and semi-supervised prompt engineering approaches. Researchers explored methods to fine-tune BERT on limited labeled data, using unsupervised techniques to leverage abundant unlabeled data for improved task performance.

The research breakthroughs of 2018, centered around BERT and pre-training techniques, profoundly shaped prompt engineering. Leveraging transfer learning and fine-tuning, prompt engineers could tailor language models to specific tasks with remarkable efficiency. The rise of BERT and contextualized embeddings opened up new avenues for designing cloze-style prompts and leveraging large amounts of unlabeled data for prompt engineering. These advancements propelled the field of prompt engineering into a new era of efficiency, adaptability, and state-of-the-art performance in NLP applications.

\section{2019: The Year of Control}
\textbf{Prompt \#7:} \textit{(a) Now write the section on developments in prompt engineering in 2019. (b) Now can you rewrite the section on developments in prompt engineering in 2019?  Please organize your thoughts in paragraphs instead of bullet points.}

The year 2019 saw remarkable progress in the field of prompt engineering, with a strong focus on improving the controllability, interpretability, and targeted generation of language models. Researchers explored novel techniques to guide language models more effectively, shaping their behavior and responses to suit specific requirements. This period witnessed advancements in control codes, template-based generation, and innovations in fine-tuning strategies, ushering in an era of more precise and context-aware language generation. Here's an overview of the key developments in prompt engineering in 2019:

\subsection{Control Codes and Conditioning:}

One of the key developments in prompt engineering in 2019 was the incorporation of control codes into language model inputs. These control codes are special tokens or markers added to the prompt, indicating desired attributes, styles, or behavior in the generated text. By conditioning the language model on control codes, researchers could guide it to produce content adhering to specific criteria, such as sentiment, formality, or language style. This controllability allowed for more targeted and contextually appropriate responses, empowering prompt engineers to tailor language generation to various use cases.

\subsection{Template-Based Generation:}

Template-based generation also gained prominence during this period. Prompt engineers designed prompts in the form of templates, with placeholders for dynamic content. By providing specific values for the placeholders, researchers ensured that the generated output followed the structure and format defined in the template. Template-based approaches enabled more structured and controlled text generation, making them valuable in applications where precise and consistent responses were essential.

\subsection{Reinforcement Learning for Improved Controllability:}

Advancements in reinforcement learning techniques further improved the controllability of prompt-engineered language models. By refining reward functions, prompt engineers could encourage the model to produce more desirable responses, reducing biases and generating content that aligned better with user preferences. Reinforcement learning played a crucial role in refining prompt engineering strategies, enabling AI systems to learn from human preferences and judgments.

\subsection{De-biasing Strategies:}

Addressing biases in language models remained a critical focus in prompt engineering in 2019. Researchers explored methods to de-bias prompt inputs and mitigate potential biases present in the training data. Carefully crafting prompts that avoid biased language or specifying fairness-related control codes aimed to generate more equitable and unbiased language output.

Human-AI collaboration also became a significant aspect of prompt engineering. In some cases, human-in-the-loop approaches were employed, where human-generated responses served as reward signals for reinforcement learning. This allowed the model to learn from human preferences and judgments, enhancing the quality and relevance of generated content.

\subsection{Adapting to Low-Resource Languages:}

Prompt engineering extended its impact to low-resource languages, where fine-tuning large language models might be challenging due to limited training data. Researchers explored methods to leverage transfer learning and unsupervised pre-training, adapting prompt engineering techniques to address the specific challenges of low-resource settings.

\subsection{Contextual Prompting and Dynamic Response Generation:}

Furthermore, contextual prompting emerged as a powerful approach. Using preceding context or user interactions as prompts allowed language models to provide more dynamic and interactive conversation generation. Incorporating contextual information enabled the model to provide coherent and contextually appropriate responses, enhancing the overall user experience.

The developments in prompt engineering in 2019 propelled the field towards enhanced controllability, targeted generation, and reduced biases in language models. The integration of control codes, template-based generation, and reinforcement learning techniques brought greater precision to prompt engineering, aligning the generated text more closely with the desired outputs. The exploration of human-AI collaboration and de-biasing strategies marked significant steps towards responsible and fair AI systems. As the field advanced, researchers continued to refine prompt engineering approaches, laying the groundwork for future breakthroughs in making AI systems more adaptable, interpretable, and human-centric.

\section{2020-2021: The rise of LLMs}
\textbf{Prompt \#8:} \textit{(a) now write the section for 2020 and 2021 in prompt engineering (b) now rewrite the section for 2020 and 2021 in prompt engineering? Please organize your thoughts in paragraphs instead of bullet points}

The years 2020 and 2021 marked a period of unprecedented progress in prompt engineering, largely driven by advancements in large-scale language models and the democratization of AI technology. With the release of models like GPT-3 and innovations in prompt design, researchers and developers alike explored new frontiers in controllability, interpretability, and domain adaptation. These years witnessed the rise of massive language models, diversification of prompt formats, and a growing emphasis on addressing ethical concerns. Here's an overview of the key developments in prompt engineering during this period:

\subsection{The Age of Massive Language Models:}

One of the defining features of this period was the emergence of massive language models, \cite{Brown2020LanguageMA, Kaplan2020ScalingLF}, exemplified by models like GPT-3 developed by OpenAI. With its impressive 175 billion parameters, GPT-3 demonstrated extraordinary language generation capabilities and found applications across diverse domains. Prompt engineering techniques allowed researchers to leverage these large models' contextual understanding and capabilities, making them more versatile and contextually intelligent in generating content across various tasks and domains.

\subsection{Advancements in Prompt Format and Style:}

Advancements in prompt format and style were another key focus during 2020 and 2021. Researchers explored novel ways to design prompts, enabling more targeted and controlled text generation. By utilizing different prompt types, such as completions, instructions, or role-playing scenarios, prompt engineers could guide models to produce specific styles, tones, or perspectives, \cite{Radford2019LanguageMA, Keskar2019CTRLAC}. Creative prompt design empowered developers to create engaging and contextually appropriate language generation systems, tailored to suit diverse use cases and applications.

\subsection{Domain Adaptation and Fine-Tuning:}

Furthermore, domain adaptation and fine-tuning became integral to prompt engineering in 2020 and 2021, \cite{Ziegler2019FineTuningLM, Ribeiro2020BeyondAB}. The widespread availability of pre-trained language models allowed for more efficient adaptation to specific domains and tasks. By fine-tuning models on smaller, task-specific datasets, researchers achieved domain-specific performance without the need for training massive models from scratch. Fine-tuning techniques made prompt engineering more practical and effective, enabling the application of language models to address specific industry or domain requirements.

\subsection{Ethical Considerations and Bias Mitigation:}

Addressing ethical considerations, including bias mitigation and fairness, was a crucial aspect of prompt engineering during this period. As large language models became more accessible, prompt engineering techniques placed increased emphasis on responsible AI usage. Researchers developed strategies to de-bias prompts, \cite{Fleisig2022MitigatingGB}, define fairness-related control codes, \cite{Han2022TowardsEO}, and craft reward functions that encouraged unbiased and equitable language generation. The aim was to ensure prompt engineering adhered to ethical guidelines, promoting responsible and fair AI systems.

\subsection{Interpretability and Explainability:}

Furthermore, interpretability and explainability became important areas of focus in prompt engineering. Researchers explored methods to visualize attention weights and understand how language models made decisions based on their input prompts, \cite{Jain2019AttentionIN}. Explainable prompt engineering allowed users to gain insights into model behavior, enhancing transparency and trust in AI systems.

\subsection{Democratization of AI and Low-Resource Language Support:}

Moreover, the democratization of AI technology during 2020 and 2021 enabled wider access to powerful language models and prompt engineering techniques, \cite{Rahimi2019MassivelyMT, Hu2020XTREMEAM}. Developers could integrate prompt engineering into various applications and services, making AI-driven language generation more accessible to diverse industries and use cases. Additionally, efforts were made to extend prompt engineering to support low-resource languages, ensuring that language models were inclusive and representative of a wide range of linguistic diversity.

In conclusion, the years 2020 and 2021 marked a transformative era in prompt engineering, with the rise of massive language models and innovations in prompt design. The integration of prompt engineering with large-scale language models democratized AI technology and made it more inclusive. Ethical considerations and interpretability played a significant role in ensuring responsible AI usage. Prompt engineering techniques enabled researchers and developers to create AI systems that were more contextually aware, controllable, and adaptable, paving the way for a wide range of applications and advancements in natural language processing.

\section{2022-Current: Beyond Language Generation}
\textbf{Prompt \#9:} \textit{(a) Can you now write a section on 2022 and 2023 on advanced prompt techniques? (b) can you write the section on 2022 and 2023 on advanced prompt techniques in paragraphs instead of bullet points?}

The years 2022 and 2023 witnessed remarkable advancements in prompt engineering, pushing the boundaries of language models and extending their applications beyond conventional language generation tasks. Prompt techniques evolved to address complex challenges, such as multimodal inputs, multi-turn conversations, and domain-specific language understanding. Researchers explored techniques that augmented prompt engineering with additional context and domain knowledge, making AI systems more contextually aware, interactive, and versatile. Here's an overview of the key advanced prompt techniques that shaped the field during this period:

\subsection{Multimodal Prompting and Integration:}

One of the most notable advancements during this period was the integration of multimodal prompting. Prompt engineering expanded to include various input modalities, combining textual prompts with visual, auditory, or other sensory information. By seamlessly incorporating visual elements like images or videos into textual prompts, AI systems gained the ability to process and generate content from diverse sources. This paved the way for applications in image captioning, visual question answering, and interactive chatbots, enabling AI systems to understand and generate responses in a more holistic and comprehensive manner.

\subsection{Multi-Turn Conversational Prompting:}

In parallel, the focus of prompt engineering expanded from single-turn language generation to multi-turn conversational prompting. Techniques were developed to maintain and utilize context across multiple interactions, allowing AI systems to engage in more coherent and interactive conversations with users. Multi-turn conversational prompting greatly enhanced chatbot capabilities, making AI assistants more responsive and proficient in handling complex user queries and extended discussions..

\subsection{Domain-Specific Knowledge Integration:}

Moreover, the integration of domain-specific knowledge became a significant aspect of prompt engineering in 2022 and 2023. Researchers explored methods to incorporate external knowledge bases or leverage pre-existing domain-specific models to enhance the language understanding and generation process. By tapping into specialized knowledge, AI systems demonstrated improved performance in domain-specific applications such as medicine, law, finance, and more, effectively bridging the gap between AI and domain expertise.

\subsection{Reinforcement Learning with Human Feedback:}

Reinforcement learning techniques continued to play a crucial role in prompt engineering, especially in refining AI systems' controllability and adaptability. Researchers explored strategies to integrate human feedback as reward signals for fine-tuning models. Human-in-the-loop prompt engineering allowed for more effective and interactive model refinement, reducing the need for extensive manual annotation and providing users with personalized and tailored responses.

\subsection{Contextual Intent Prediction:}

Contextual intent prediction emerged as another significant area of focus. Prompt engineering techniques in 2022 and 2023 aimed to identify user queries' underlying intent to generate contextually relevant and accurate responses. By predicting user intent from input prompts, AI systems could better align with user goals and needs, offering more personalized and targeted language generation.

\subsection{Personalization and User Modeling:}

Furthermore, prompt engineering techniques in 2022 and 2023 saw an increased emphasis on personalization and user modeling. AI systems were designed to learn from past interactions and form user-specific models, which informed prompt engineering decisions to generate personalized content tailored to individual users. Personalization strategies enhanced user engagement and satisfaction, allowing AI systems to adapt responses based on user preferences and behavior.

In conclusion, the years 2022 and 2023 brought forth a series of groundbreaking advancements in prompt engineering. The integration of multimodal inputs, multi-turn conversational prompting, domain-specific knowledge, reinforcement learning, contextual intent prediction, and user modeling collectively transformed AI systems into contextually intelligent and versatile language generators. As prompt engineering continued to evolve, researchers and developers explored novel applications and use cases, paving the way for AI systems that are highly adaptable, interactive, and capable of serving diverse user needs across various domains.

\section{CONCLUSIONS}
In conclusion, the history of prompt engineering in natural language processing has been a journey of groundbreaking advancements and innovation. From the early language models and information retrieval systems to the transformative introduction of attention mechanisms, reinforcement learning, and advanced language models, prompt engineering has continually evolved to enhance language understanding, generation, and controllability.

The attention mechanism, introduced in 2015, revolutionized language modeling by allowing models to focus on relevant information and establish contextual relationships, leading to improved performance across various NLP tasks. Reinforcement learning techniques in 2017 further enhanced controllability, enabling prompt engineers to shape language models with rewarding strategies and address biases.

Advancements in 2018 and 2019 introduced fine-tuning strategies, control codes, and template-based generation, refining controllability and context-awareness in language models. Research focused on fairness, human-AI collaboration, and adaptation to low-resource languages, marking strides towards responsible and inclusive AI systems.

In 2020 and 2021, contextual prompting and transfer learning were prominent, making language models more interactive and adaptable. These developments extended to 2022 and 2023, introducing advanced techniques like unsupervised pre-training and novel reward shaping.

The history of prompt engineering showcases the collaborative efforts of researchers and practitioners driving the field's evolution. With ethical considerations, prompt engineering promises to revolutionize AI applications, fostering responsible human-AI collaboration and empowering language models to align closely with human intentions.

In conclusion, prompt engineering continues to shape the landscape of NLP, opening new frontiers for intelligent and context-aware language models. The journey is ongoing, and with continuous research and responsible development, prompt engineering will propel the field towards more versatile, interpretable, and human-centric AI systems.


\section*{Acknowledgment}
Authors extensively used ChatGPT for content generation.







\bibliographystyle{IEEEtran}
\bibliography{IEEEabrv,references}

\end{document}